\def\year{2021}\relax
\documentclass[letterpaper]{article} 
\usepackage{aaai21}  
\usepackage{times}  
\usepackage{helvet} 
\usepackage{courier}  
\usepackage[hyphens]{url}  
\usepackage{graphicx} 
\usepackage{natbib}  
\usepackage{caption} 
\usepackage{url}            
\usepackage{booktabs}       
\usepackage{amsmath, amsfonts, amsthm}

\usepackage{soul}
\usepackage{subcaption}
\usepackage{tabularx, multirow}
\usepackage{bm}
\usepackage{dsfont}
\usepackage[ruled,vlined]{algorithm2e}
\usepackage{algpseudocode}
\usepackage{leftidx}

\newcommand{\cam}{CAM\xspace}

\newcommand{\gap}{GAP\xspace}
\newcommand{\gmp}{GMP\xspace}

\newcommand{\fcn}{FCN\xspace}
\newcommand{\resnet}{ResNet\xspace}

\newcommand{\proposed}{DTP\xspace}
\newcommand{\dtp}{DTP\xspace}
\newcommand{\gtp}{GTP\xspace}
\newcommand{\stp}{STP\xspace}

\newcommand{\xseries}{\mathbf{X}}
\newcommand{\hseries}{\mathbf{H}}
\newcommand{\gseries}{\mathbf{\bar{H}}}
\newcommand{\pseries}{\mathbf{P}}
\newcommand{\wseries}[1]{\mathbf{W}^{(#1)}}

\newcommand{\xvec}[1]{\mathbf{x}_{#1}}
\newcommand{\hvec}[1]{\mathbf{h}_{#1}}
\newcommand{\gvec}[1]{\mathbf{\bar{h}}_{#1}}
\newcommand{\pvec}[1]{\mathbf{p}_{#1}}
\newcommand{\wvec}[2]{\mathbf{w}_{#1}^{(#2)}}

\newcommand{\sdtw}[1]{\text{DTW}_{#1}}
\newcommand{\dtw}{\text{DTW}}
\newcommand{\closs}{\mathcal{L}_{class}}
\newcommand{\ploss}{\mathcal{L}_{proto}}

\DeclareMathOperator*{\argmin}{argmin}

\title{Learnable Dynamic Temporal Pooling for Time Series Classification}

\author{Dongha Lee\textsuperscript{\rm 1}, Seonghyeon Lee\textsuperscript{\rm 2}, Hwanjo Yu\textsuperscript{\rm 2}\thanks{Corresponding author}\\}
\affiliations{
\textsuperscript{\rm 1}Institute of Artificial Intelligence, POSTECH, Republic of Korea\\
\textsuperscript{\rm 2}Dept. of Computer Science and Engineering, POSTECH, Republic of Korea\\
\{dongha.lee, sh0416, hwanjoyu\}@postech.ac.kr
}

\begin{document}

\maketitle
\begin{abstract}
With the increase of available time series data, predicting their class labels has been one of the most important challenges in a wide range of disciplines.
Recent studies on time series classification show that convolutional neural networks (CNN) achieved the state-of-the-art performance as a single classifier.
In this work, pointing out that the global pooling layer that is usually adopted by existing CNN classifiers discards the temporal information of high-level features, we present a dynamic temporal pooling (\dtp) technique that reduces the temporal size of hidden representations by aggregating the features at the segment-level.
For the partition of a whole series into multiple segments, we utilize dynamic time warping (DTW) to align each time point in a temporal order with the prototypical features of the segments, which can be optimized simultaneously with the network parameters of CNN classifiers.
The \dtp layer combined with a fully-connected layer helps to extract further discriminative features considering their temporal position within an input time series.
Extensive experiments on both univariate and multivariate time series datasets show that our proposed pooling significantly improves the classification performance.
\end{abstract}

\section{Introduction}
\label{sec:intro}
In the last two decades, time series classification has been dominated by nearest neighbor classifiers which utilize handcrafted feature-based representations~\cite{baydogan2013bag,schafer2015boss} or various distance measures between time series~\cite{zhao2018shapedtw,yuan2019locally}.
Recently, there have been several attempts to exploit deep neural networks (DNN) for time series classifiers~\cite{zheng2016exploiting,zhao2017convolutional,wang2017time};
they do not require heavy crafting on feature engineering or data preprocessing, and easily be applied to multivariate time series as well.
In practice, fully convolutional networks (FCN) and residual networks (ResNet) designed for time series classification~\cite{wang2017time} showed the state-of-the-art accuracy among various DNN competitors~\cite{fawaz2019deep}.

However, the existing convolutional neural networks (CNN) are not able to fully utilize the temporal information of high-level features for their classification.
On top of convolutional layers, the CNN classifiers adopt global average pooling (GAP) or global max pooling (GMP) that simply aggregates all hidden vectors along the time axis.
By doing so, they can obtain a global representation for an input time series, as well as avoid the overfitting problem with the help of much fewer model parameters.
Nevertheless, such global aggregation discards the temporal position of the hidden features, which makes the CNN learn only position-invariant temporal features.
Since each temporal position itself could be a useful feature for discrimination among different classes in time series classification, the global pooling layer eventually degrades the performance of the classifiers.

To address this limitation, we propose a novel pooling method that effectively reduces the temporal size (i.e., length) of network outputs while minimizing the loss of temporal information.
Motivated by the observations that time series instances consist of multiple segments with distinct patterns, our dynamic temporal pooling (\dtp) outputs a pooled vector for each segment rather than the one for a whole series.
The \dtp layer produces segment-level representations by aggregating hidden vectors in each segment, thus it enables to model the classification score based on segment-specific class weights.
In other words, our CNN classifiers replace the global pooling with the segment-level pooling (being followed by a fully-connected layer), which allows extracting further class-discriminative features and improves the classification accuracy.
We additionally present the class activation map (CAM) specifically for our \dtp layer, indicating how much each temporal region contributes to predicting the class label of an input time series.

\begin{figure*}[t]
	\centering
	\begin{subfigure}{0.31\linewidth}
	\centering
	\includegraphics[width=0.95\linewidth]{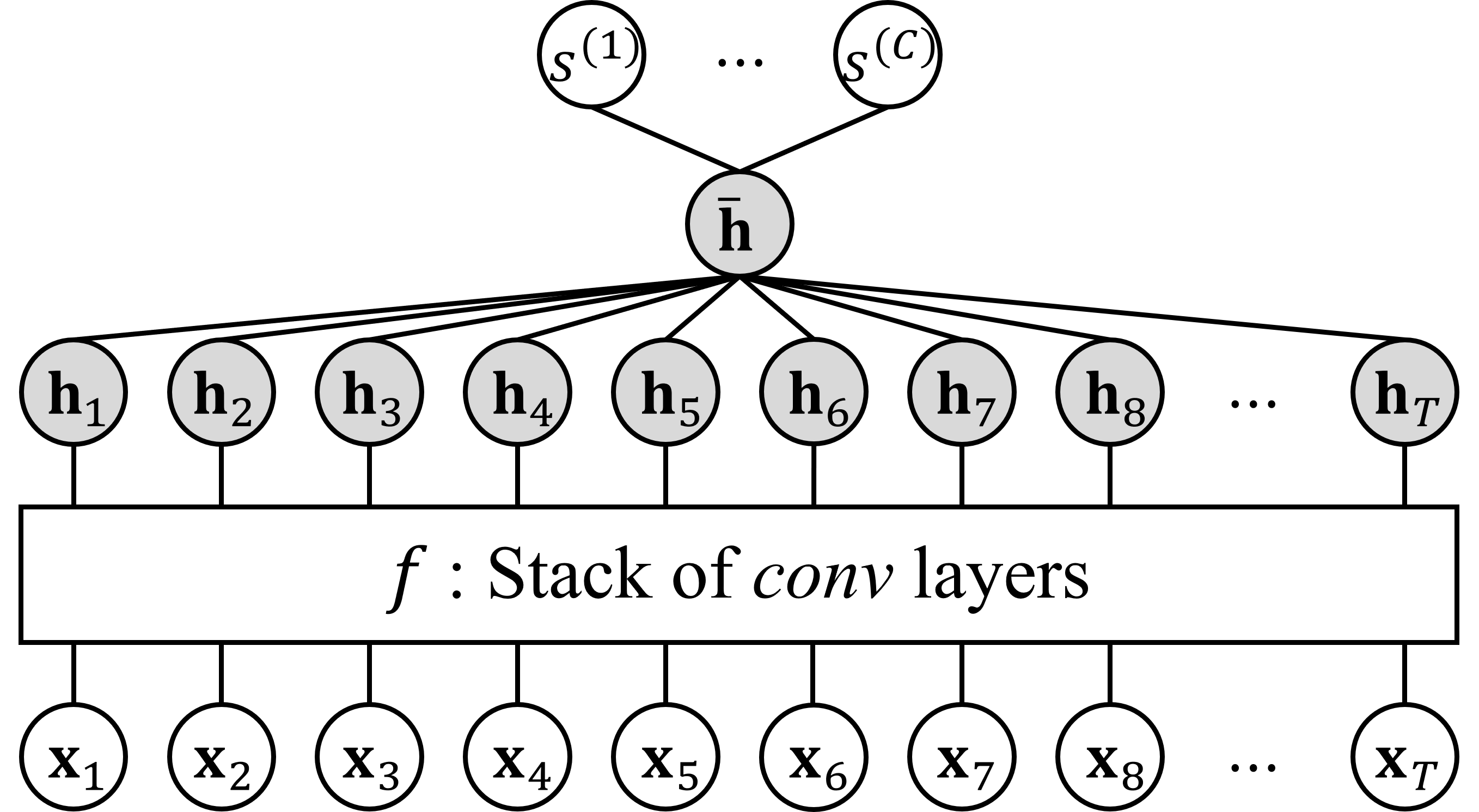}
	\caption{Global temporal pooling}
	\label{fig:gtp}
	\end{subfigure}
	\begin{subfigure}{0.31\linewidth}
	\centering
	\includegraphics[width=0.95\linewidth]{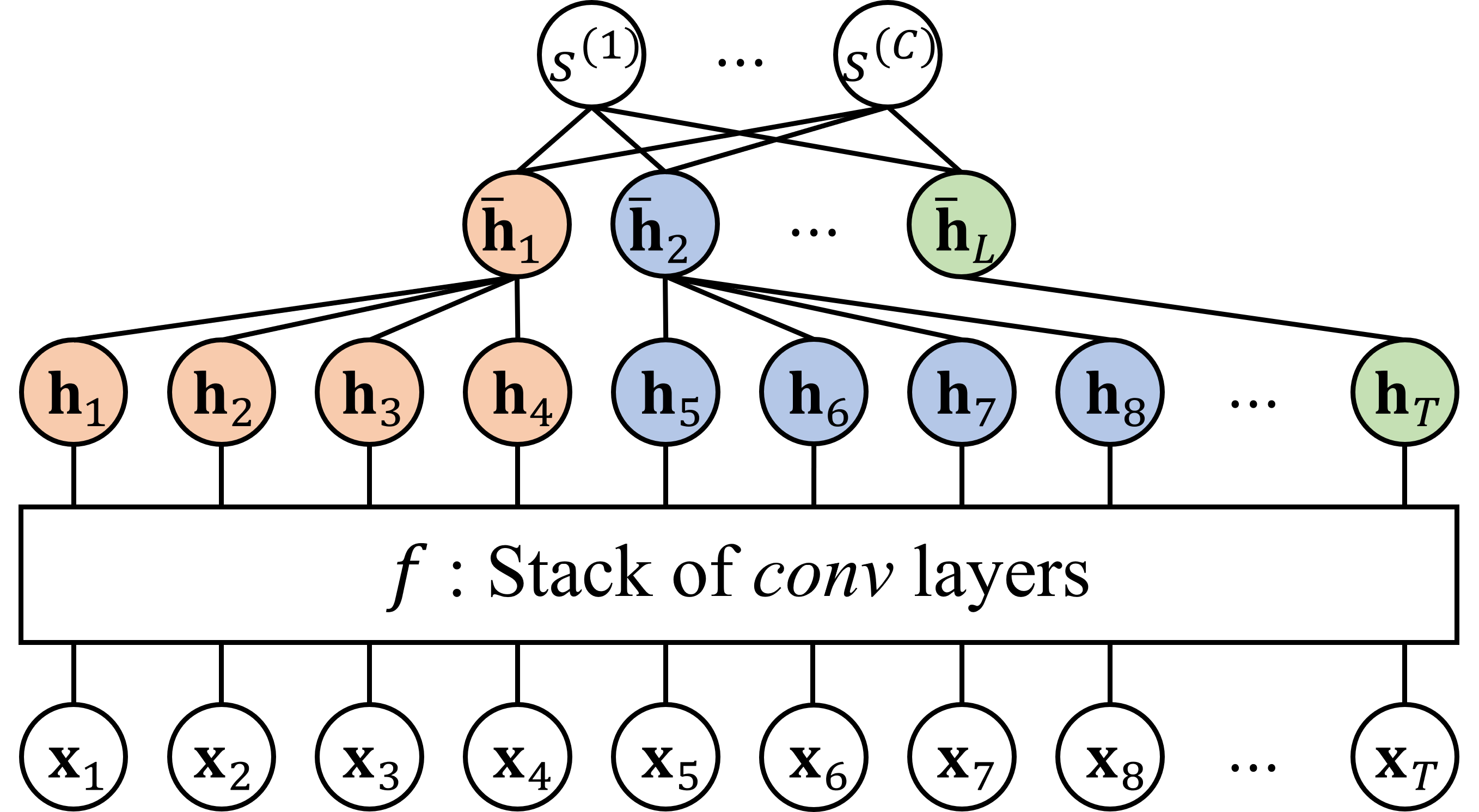}
	\caption{Static temporal pooling}
	\label{fig:stp}
	\end{subfigure}
	\begin{subfigure}{0.31\linewidth}
	\centering
	\includegraphics[width=0.95\linewidth]{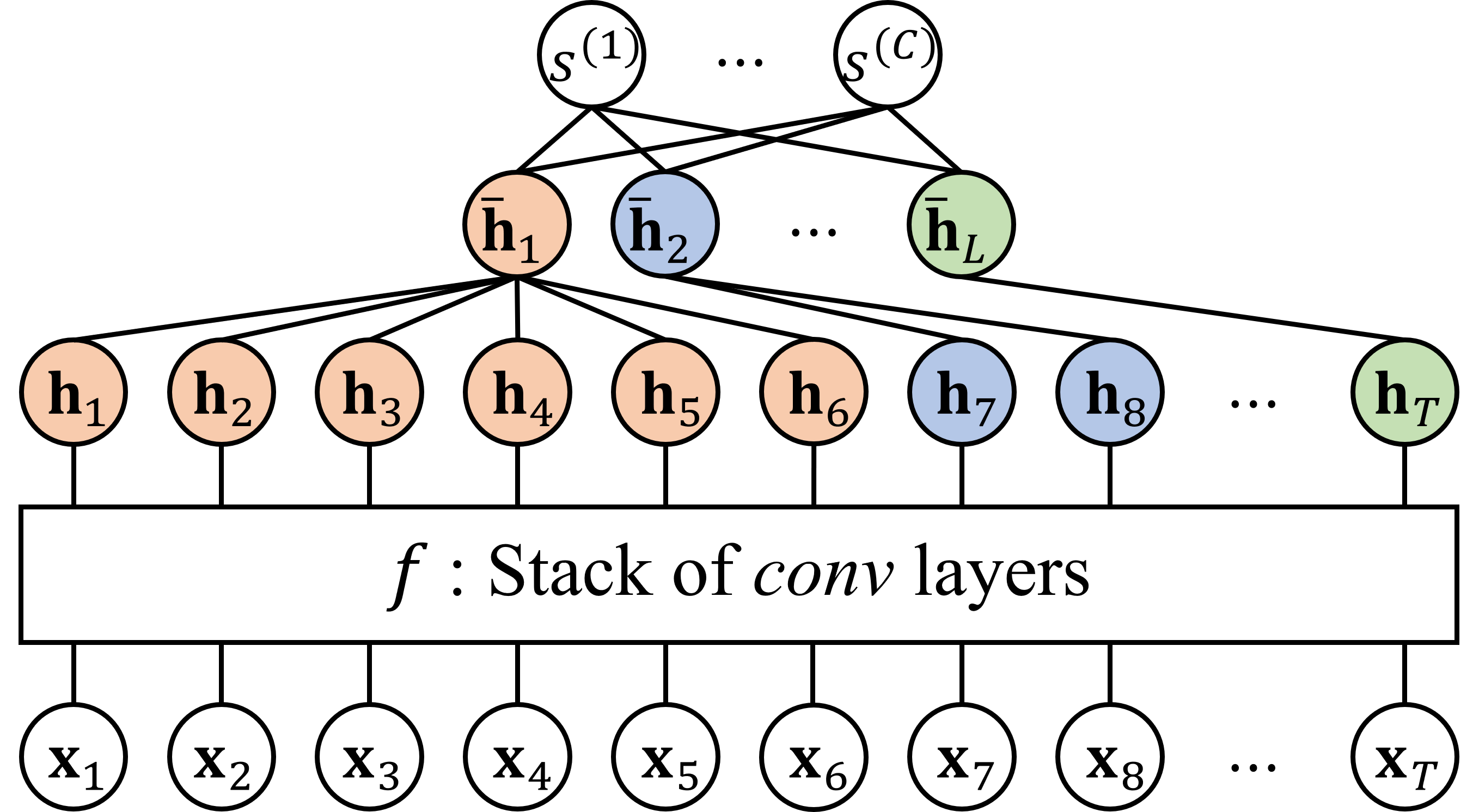}
	\caption{Dynamic temporal pooling}
	\label{fig:dtp}
	\end{subfigure}
	\caption{The architectures of CNN classifiers with different temporal pooling layers. 
	The global pooling simply aggregates all the hidden vectors, while the proposed temporal pooling effectively reduces the temporal size based on time series segmentation.}
	\label{fig:framework}
\end{figure*}

The challenge here is to find out consistent segments from input time series instances that are not temporally aligned with each other.
To this end, the \dtp layer performs semantic segmentation by using dynamic time warping (DTW).
We first introduce trainable latent vectors as many as the number of segments to be identified, termed as \textit{prototypical hidden series}, for encoding the prototypical features of each segment into them in a temporal order.
Then, the \dtp layer aligns the network outputs (i.e., the series of hidden vectors) with the prototypical hidden series while keeping their temporal order based on DTW; this generates the set of consecutive time points matching with each segment.
In the end, we simultaneously optimize the network parameters of CNN classifiers and the prototypical hidden series, thereby both of them collaboratively improve with each other.
That is, training the CNN assists to capture the prototypical features of the segments, and also learning the prototypical hidden series helps the CNN to extract discriminative features.

Our empirical evaluation on extensive univariate and multivariate time series datasets demonstrates that the proposed pooling technique significantly improves the discrimination power of the CNN classifiers, regardless of its network architecture or pooling operation.
The CNN classifiers with the \dtp layer also beat nearest neighbor classifiers that use various distance measures by a large margin.
Furthermore, we qualitatively show that the \dtp layer is capable of providing an interpretable analysis on its classification result by localizing discriminative regions within a target time series.

\section{Related Work}
\label{sec:related}
\subsection{Deep Learning for Time Series Classification}
\label{subsec:dltsc}
With the great success of deep learning, a variety of deep neural networks (DNN) have been applied to time series classification, and they achieved higher accuracy than conventional nearest neighbor classifiers~\cite{fawaz2019deep}.
Among them, convolutional neural networks (CNN) have gained much attention, such as fully convolutional networks (\fcn) and residual networks (\resnet)~\cite{wang2017time},
because of their capability of capturing local patterns as well as efficient inference by parallel computations.
As presented in Figure~\ref{fig:framework}, the stack of multiple convolutional layers outputs the hidden vector at each time point,\footnote{The CNN architectures, we focus on in this work, do not use any local pooling layers, so they keep the temporal size (length) of hidden representations unchanged throughout the convolutions.}
which eventually encodes \textit{high-level} features about the local temporal context\footnote{In a univariate case, the local contexts correspond to shapelets \cite{lines2012shapelet,ma2020adversarial} of the receptive field size.} surrounding the time point.
Based on global average pooling (\gap) or global max pooling (\gmp), all the hidden vectors are summarized along the time axis into a single vector (Figure~\ref{fig:gtp}), and it is finally used for computing classification scores.

However, such global pooling causes the loss of information about temporal dynamics of the high-level features, and this leads to the limited performance.
In case of time series classification, local temporal patterns can have different meanings depending on their temporal positions where they occur (i.e., position-variant), unlike image classification where the position of visual semantic features does not much affect its class label (i.e., position-invariant).
Nevertheless, using the fully-connected layer (with local pooling after each convolution) instead of the global pooling layer, e.g., Time-LeNet~\cite{le2016data}, not only requires a large number of parameters increasing with the length of time series but also makes the network overfitted to the training data~\cite{fawaz2019deep}.

\subsection{Differentiable Dynamic Time Warping}
\label{subsec:softdtw}
Dynamic time warping (DTW) is a popular technique for measuring the distance between two time series of different lengths, based on point-to-point matching with the temporal consistency.
Given two times series $X$ and $Y$ of length $M$ and $N$, the $(m, n)$-th entry of its cost matrix $\Delta(X, Y)\in\mathbb{R}^{M\times N}$ represents the distance (or alignment cost) between $X_m$ and $Y_n$.
The DTW distance between $X$ and $Y$ is defined by the minimum inner product of the cost matrix and any binary alignment matrix $A$,
\begin{equation}
\label{eq:dtw}
\small
    \dtw(X, Y) =  \text{min}\left\{ \langle A, \Delta(X, Y)\rangle, \forall A\in \mathcal{A} \right\},
\end{equation}
where $\mathcal{A}\subset \{0, 1\}^{M \times N}$ is the set of possible binary alignment matrices whose $(m, n)$-th entry indicates whether $X_m$ and $Y_n$ are aligned or not.
Each alignment matrix corresponds to a warping path that connects the upper-left $(1, 1)$-th entry to the lower-right $(L, T)$-th entry using $\downarrow, \rightarrow, \searrow$ moves.
That is, DTW searches for the optimal warping path that minimizes the total alignment cost, and the path eventually represents the best temporal alignment between the two series.
The DTW distance and its alignment matrix can be efficiently obtained by dynamic programming based on Bellman recursion, which takes a quadratic $O(MN)$ cost.

Recently, the continuous relaxation of DTW~\cite{cuturi2017soft}, named as soft-DTW, has been proposed in order to calculate the gradient of DTW with respect to its input series.
Instead of the discontinuous (i.e., non-differentiable) hard-min operation taking only the minimum value in Equation~\eqref{eq:dtw}, soft-DTW utilizes the soft-min operation with a smoothing parameter $\gamma$ by adopting the concept of global alignment kernels~\cite{cuturi2007kernel}.
The soft-DTW distance between $X$ and $Y$ is calculated by
\begin{equation}
\label{eq:softdtw}
\begin{split}
\small
    \sdtw{\gamma}(X, Y) &=  \text{min}_{\gamma}\left\{ \langle A, \Delta(X, Y)\rangle, \forall A\in \mathcal{A} \right\}, \\
    \text{min}_{\gamma} \{a_1, \ldots, a_n\} &=
    \begin{cases}
    \; \min_{i\leq n} a_i, & \gamma = 0\\
    \; -\gamma \log \sum_{i=1}^{n}e^{-a_i/\gamma}, & \gamma > 0.
    \end{cases}
\end{split}
\end{equation}
Note that the original DTW is the special case of soft-DTW with $\gamma=0$.
The differentiability of soft-DTW facilitates the optimization of distance (or similarity) among time series in deep learning frameworks.

\section{Dynamic Temporal Pooling}
\label{sec:proposed}

\subsection{Problem Formulation}
Given a training set of $N$ time series instances from $C$ classes, $\{(\xseries^1, y^1), \ldots, (\xseries^N, y^N)\}$ where $y \in \{1, \ldots, C\}$,
we aim to train a CNN classifier that accurately predicts the class label of an input time series instance. 
For each instance $\xseries = [\xvec{1}, \ldots, \xvec{T}] \in \mathbb{R}^{D \times T}$ of length $T$ with $D$ variables, the network $f$ parameterized by $\mathcal{W}$ outputs the series of hidden vectors $\hseries = [\hvec{1}, \ldots, \hvec{T}] \in \mathbb{R}^{K \times T}$ of dimension size $K$.\footnote{For notational convenience, we omit the superscript $n$ that represents the instance index in the rest of the paper.}
In the end, the final classification scores $\mathbf{s}=[s^{(1)},\ldots, s^{(C)}]$ are
computed from the hidden vectors.

\subsection{Temporal Pooling based on Segmentation}
\label{subsec:dtp}
The purpose of our temporal pooling is to reduce the temporal size $T$ of the hidden representation (i.e., the output of $f$), while minimizing the loss of temporal information in the time series.
The key idea is to partition a series of hidden vectors into $L$ segments ($L \ll T$) then generate pooled representations by summarizing the vectors in each segment.
Formally, the temporal pooling layer outputs the series of pooled vectors $\gseries = [\gvec{1}, \ldots, \gvec{L}] \in \mathbb{R}^{K \times L}$ of length $L$, whose $l$-th vector is obtained as follows.
\begin{equation}
\label{eq:tp}
\small
    \gvec{l} = \phi\left(\hvec{t_{l-1}+1}, \hvec{t_{l-1}+2}\ldots, \hvec{t_{l}}\right),
\end{equation}
where $\phi$ is the pooling operation, and $\mathcal{T}_l=\{t_{l-1}+1, \ldots, t_{l}\}$ is the set of consecutive time points belonging to the $l$-th segment ($t_0=0, t_L=T$).
Three functions can be used for the pooling operation:
calculating the average value (denoted by \textit{avg}),
the summation value (denoted by \textit{sum}),
and the maximum value (denoted by \textit{max}) for each latent dimension.

\begin{figure}[t]
	\centering
	\includegraphics[width=\linewidth]{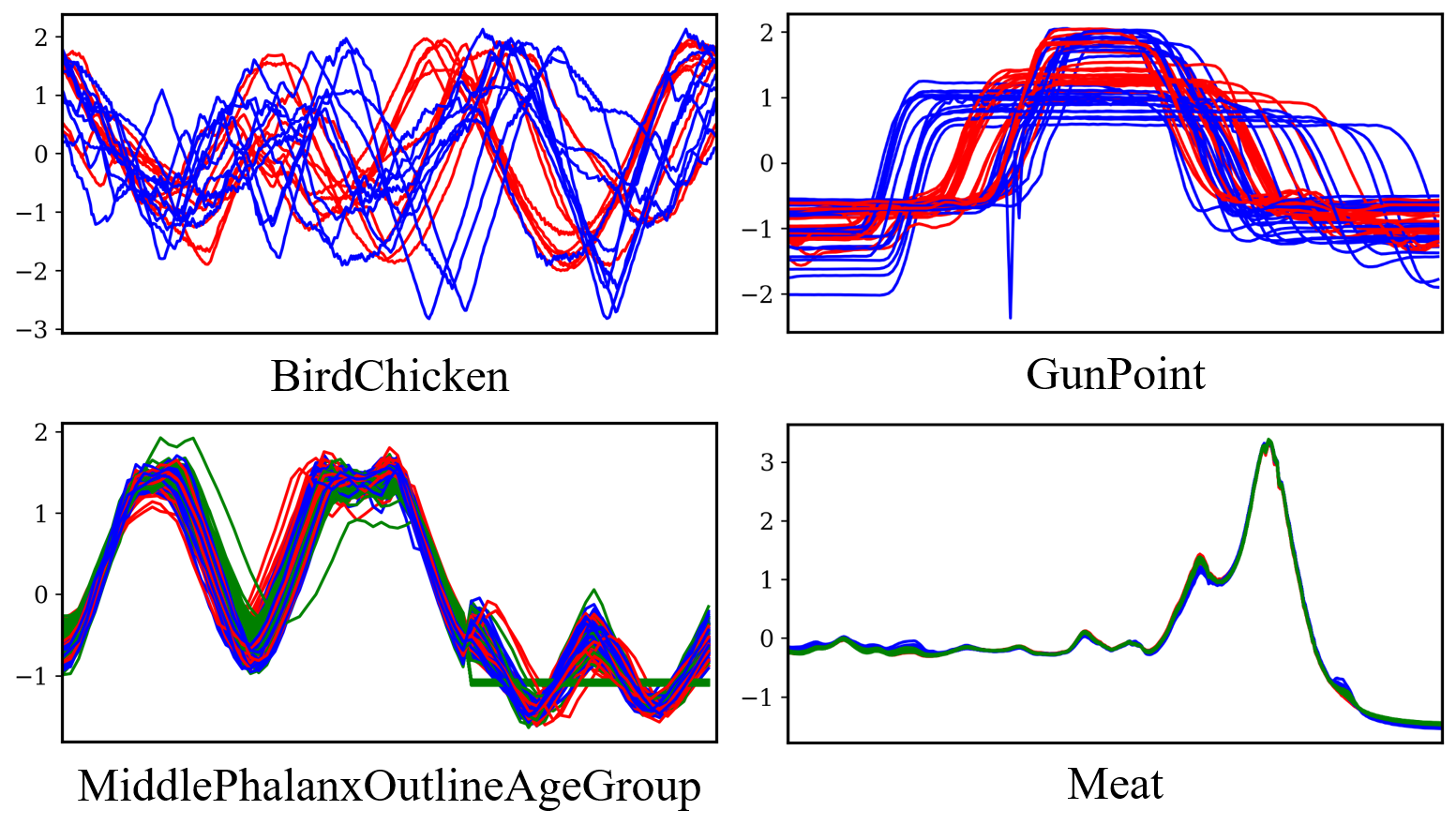}
	\caption{Examples of univariate time series, whose classes are marked in different colors.
	(Best viewed in color.)}
	\label{fig:patterns}
\end{figure}

A straightforward strategy for segmentation is to partition a whole time series into shorter ones of the same length in a static manner (Figure~\ref{fig:stp}), but it has several limitations that have to be addressed.
First, time points from different time series instances are not temporally aligned in general, which makes it difficult to find absolute temporal locations for segmentation (Figure~\ref{fig:patterns}, upper).
Furthermore, considering a segmentation task aims to discover distinct temporal patterns which are internally homogeneous, the length of each \textit{optimal} time series segment cannot be the same as the others' in most cases (Figure~\ref{fig:patterns}, lower).

\begin{figure}[t]
	\centering
	\includegraphics[width=\linewidth]{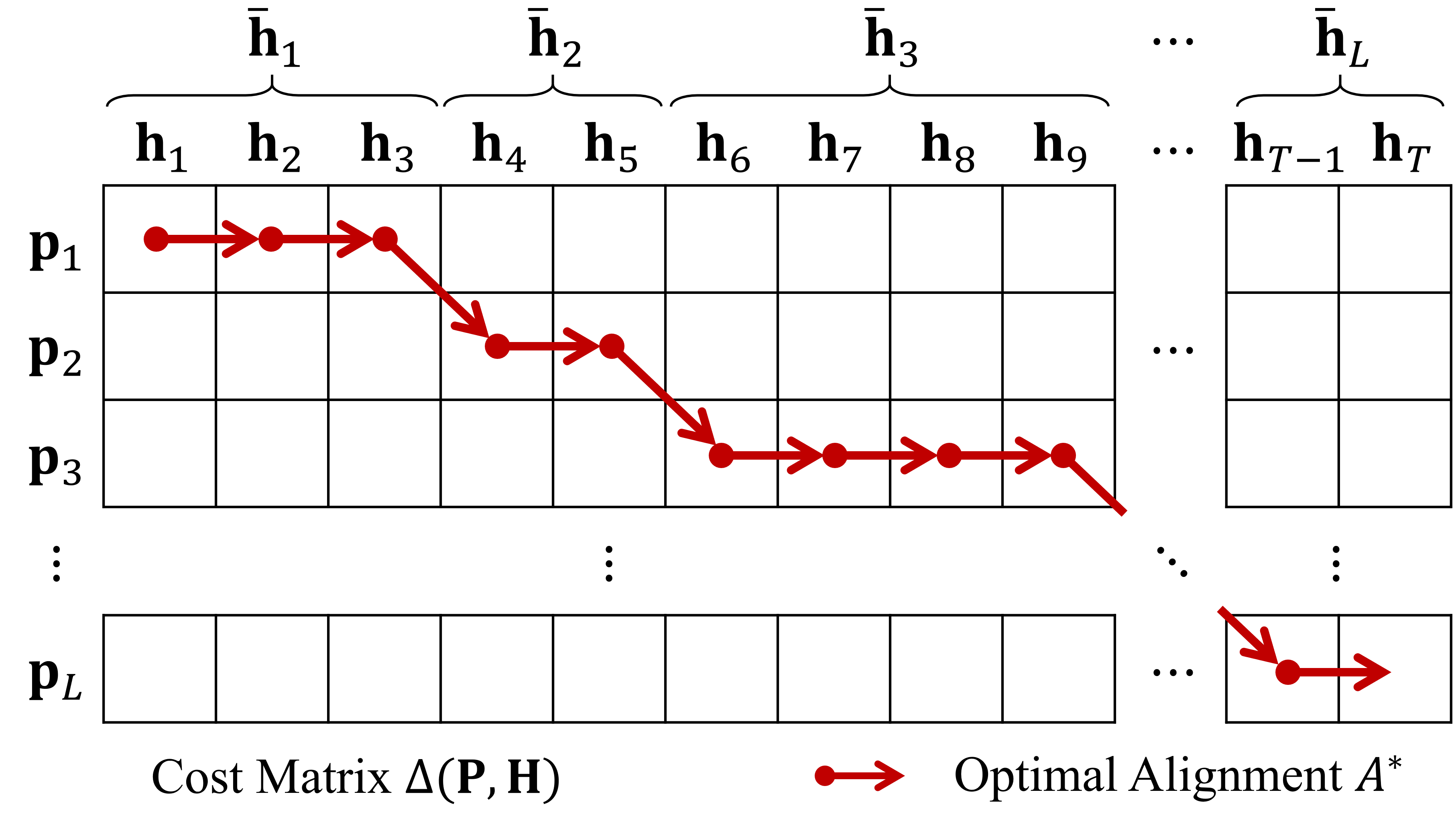}
	\caption{\proposed finds $L$ segments from the series of hidden vectors of length $T$ based on DTW alignment, then summarizes the vectors within each segment.}
	\label{fig:alignment}
\end{figure}

To tackle these challenges, we perform semantic segmentation by matching each time point with its semantically-closest segment in a temporal order.
To this end, we first introduce a \textit{prototypical hidden series} $\pseries=[\pvec{1},\ldots, \pvec{L}]\in\mathbb{R}^{K\times L}$ of length $L$ which best summarizes the high-level features of $L$ segments. 
Our proposed method, termed as dynamic temporal pooling (\dtp), temporally aligns the series of prototypical hidden vectors (i.e., $\pseries$) with that of target hidden vectors (i.e., $\hseries$) by using DTW.
Based on the result of DTW alignment, the series of hidden vectors is partitioned into $L$ segments (Figure~\ref{fig:alignment}), and each of them is pooled by Equation~\eqref{eq:tp}.
In this case, the optimal alignment matrix $A^*$ between $\pseries$ and $\hseries$ can be obtained by
\begin{equation}
\label{eq:alignment}
\small
    A^* = \argmin_{A\in\mathcal{A}}\ \langle A, \Delta(\pseries, \hseries)\rangle.
\end{equation}
$\Delta\in\mathbb{R}^{L\times T}$ is the alignment cost matrix whose $(l, t)$-th entry $\delta(\pvec{l}, \hvec{t})$ encodes the distance between $\pvec{l}$ and $\hvec{t}$.
We define the cost by using their cosine similarity as follows.\footnote{We also consider the Euclidean distance $\lVert\pvec{l}-\hvec{t}\rVert_2$ and the dot product $\exp(-\pvec{l}\cdot\hvec{t})$, but we empirically found that the cosine distance shows the best performance among them.}
\begin{equation}
\label{eq:costfunc}
\small
\delta(\pvec{l}, \hvec{t}) = 1 - \frac{\pvec{l}\cdot\hvec{t}}{\lVert\pvec{l}\rVert_2\lVert\hvec{t}\rVert_2}.
\end{equation}

Unlike eligible warping paths of conventional DTW, we need to impose an additional constraint that each time point should be aligned only with a single segment.
Thus, we limit $\mathcal{A}$ to the set of possible binary alignment matrices representing a path that connects the upper-left $(1, 1)$-th entry to the lower-right $(L, T)$-th entry using only $\rightarrow, \searrow$ moves.

\subsection{Learnable Dynamic Temporal Pooling Layer}
\subsubsection{Learning the prototypical hidden series}
\label{subsec:protos}
For effective segmentation, the prototypical hidden series $\pseries$ should be optimized so that its $l$-th vector learns the latent semantic (or prototypical features) corresponding to the segment $l$.
Given a training set, we obtain the prototypical hidden series by minimizing its soft-DTW distance from the hidden representations of all the time series instances.
\begin{equation}
\label{eq:ploss}
\small
    \ploss(\pseries) = \frac{1}{N}\sum_{n=1}^{N} \sdtw{\gamma}(\pseries, f(\xseries^{n};\mathcal{W})) .
\end{equation}
As we discussed, soft-DTW is differentiable with respect to its input, thus $\pseries$ can be easily optimized using the gradients $\nabla_{\pseries} \sdtw{\gamma}$ produced by Equation~\eqref{eq:ploss}.

Similar to the original DTW, the soft-DTW distance (and the soft alignment matrix) can be obtained by solving a dynamic program based on Bellman recursion.
Algorithm~\ref{alg:softdtw} describes the process of forward and backward recursions to compute the alignment cost $\sdtw{\gamma}(\pseries, \hseries)$ and its gradient $\nabla_{\pseries} \sdtw{\gamma}(\pseries, \hseries)$.  
Please refer to~\cite{cuturi2017soft} for more details of deriving the algorithms.
Note that a single time point should not be aligned with multiple consecutive segments.
For this reason, our forward recursion does not allow the $\downarrow$ relation in its recurrence (i.e., $R_{l,t}$ depends on only $R_{l-1,t-1}$ and $R_{l,t-1}$), 
and accordingly, the backward recursion does not consider the $\uparrow$ relation (i.e., $E_{l,t}$ is obtained from $E_{l,t+1}$ and $E_{l+1,t+1}$).

\begin{algorithm}[t]
    \small
    \DontPrintSemicolon
    \SetKwProg{Fn}{Function}{:}{}
    \SetKwComment{Comment}{$\triangleright$\ }{}
    \Fn{$\text{forward} \left(\pseries, \hseries \right)$}{
    \Comment*[r]{\rmfamily Fill the alignment cost matrix $R\in\mathbb{R}^{L\times T}$}
    $R_{0,0} = 0$, 
    $R_{:,0} = R_{0,:} = \infty$ \; 
    \For{$l=1,\ldots,L$}{
        \For{$t=1,\ldots,T$}{
            $R_{l,t} = \delta(\pvec{l}, \hvec{t}) + \text{min}_\gamma\{R_{l-1,t-1}, R_{l,t-1}\}$
        }
    }
    \Return $\sdtw{\gamma}(\pseries, \hseries) = R_{L,T}$
    }
    \Fn{$\text{backward} \left(\pseries, \hseries \right)$}{
    \Comment*[r]{\rmfamily Fill the soft alignment matrix $E\in\mathbb{R}^{L\times T}$}
    $E_{L+1,T+1} = 1$ \Comment*[r]{\rmfamily $E_{l,t}:=\partial R_{L,T}/\partial R_{l,t}$}
    $E_{:,T+1} = E_{L+1,:} = 0$ \;
    $R_{:,T+1} = R_{L+1,:} = -\infty$ \;
    \For{$l=L,\ldots,1$}{
        \For{$t=T,\ldots,1$}{
        $a = \exp\frac{1}{\gamma}(R_{l,t+1}-R_{l,t}-\delta(\pvec{l},\hvec{t+1}))$ \;
        $b = \exp\frac{1}{\gamma}(R_{l+1,t+1}-R_{l,t}-\delta(\pvec{l+1},\hvec{t+1}))$ \;
        \vspace{-10pt}
        $E_{l,t} = a \cdot E_{l,t+1} + b \cdot E_{l+1,t+1}$
        }
    }
    \vspace{-5pt}
    \Return $\nabla_{\pseries}\sdtw{\gamma}(\pseries, \hseries) = \left(\frac{\partial \Delta(\pseries, \hseries)}{\partial \pseries}\right)^T E$
    }
	\caption{Forward and backward recursions to compute $\sdtw{\gamma}(\pseries, \hseries)$ and  $\nabla_{\pseries}\sdtw{\gamma}(\pseries, \hseries)$}
	\label{alg:softdtw}
\end{algorithm}

\subsubsection{Learning the parameters of the CNN classifier}
\label{subsec:clf}
Instead of the class weight vector $\mathbf{w}^{(c)}\in\mathbb{R}^{K}$ that learns the importance of each latent dimension for class $c$, we introduce a class weight matrix $\wseries{c} = [\wvec{1}{c}, \ldots, \wvec{L}{c}] \in \mathbb{R}^{K \times L}$ in order that the class weights are independently modeled for each segment.
Using the pooled vectors and the class weight matrices, we calculate the classification score $s^{(c)} = \sum_{l=1}^{L} \gvec{l} \cdot \wvec{l}{c}$, and the posterior probability that an input time series instance belongs to class $c$ is defined as follows:
\begin{equation}
\label{eq:posterior}
\small
    P(y=c|\xseries) = \frac{\exp \left(\sum_{l=1}^{L}\gvec{l} \cdot \wvec{l}{c}\right) }{\sum_{c'=1}^{C} \exp \left(\sum_{l=1}^{L}\gvec{l} \cdot \wvec{l}{c'}\right)},
\end{equation}
The network parameters $\mathcal{W}$ and class weight matrices $\{\wseries{c}\}$ are optimized by the following classification loss.
\begin{equation}
\label{eq:closs}
\small
    \closs(\mathcal{W}, \{\wseries{c}\}) = -\frac{1}{N}\sum_{n=1}^{N}\log P(y=y^n|\xseries^n)
\end{equation}
To sum up, our CNN classifier adopts the segment-level fully-connected layer for time series classification, by combining Equation~\eqref{eq:posterior} with the \dtp layer.

\subsubsection{Optimization}
All the parameters including the prototypical hidden series, the network parameters, and the class weight matrices are effectively optimized at the same time, by minimizing the corresponding losses:
\begin{equation}
\begin{split}
\small
    &\pseries \leftarrow \pseries-\eta\cdot {\partial \ploss}/{\partial \pseries}, \\
    &\mathcal{W} \leftarrow \mathcal{W}-\eta\cdot {\partial \closs}/{\partial \mathcal{W}}, \\
    &\mathbf{W}^{(c)} \leftarrow \mathbf{W}^{(c)}-\eta\cdot {\partial \closs}/{\partial \mathbf{W}^{(c)}}.
\end{split}
\end{equation}

Note that we consider $\hseries=f(\xseries;\mathcal{W})$ rather than $\xseries$ for DTW alignment (Equations~\eqref{eq:alignment} and \eqref{eq:ploss}), so as to perform semantic segmentation in the latent space where the high-level features are embedded. 
For this reason, $\pseries$ and $\hseries$ collaboratively improve with each other during the training, and each of them respectively captures \textit{stereotypical} macro-patterns and \textit{discriminative} micro-ones.
Training the network parameters of CNN classifiers (Equation~\eqref{eq:closs}) helps to learn the prototypical high-level features of each segment, 
while optimizing the prototypical hidden series (Equation~\eqref{eq:ploss}) facilitates the CNN to learn further discriminative features by providing more consistent segments. 

\subsection{Class Activation Map for Temporal Pooling Layer}
\label{subsec:dtpcam}
For further explanation on how much each temporal region (or time point) contributes to the classification of a target time series, we tailor a class activation map (\cam)~\cite{zhou2016learning,wang2017time} for CNN classifiers with the \dtp layer.
CAM has been used to localize the regions relevant to the target class within a time series, but it cannot take into account the segment-specific class weights and also requires the class weights trained for the GAP layer. 

Inspired by Grad-CAM~\cite{selvaraju2017grad}, we generalize the activation map for class $c$ at time point $t$ by
\begin{equation}
\label{eq:dtpcam}
\small
    M_t^{(c)}= \sum_{k=1}^{K} \left(\frac{\partial s^{(c)}}{\partial h_{t, k}} \cdot h_{t,k} \right).
\end{equation}
The point-wise gradient $\partial s^{(c)}/\partial h_{t,k}$, which implies the position-specific class weights, is used as the weight for each hidden feature $h_{t,k}$. 
For example, $M_t^{(c)}$ becomes equivalent to $\wvec{l}{c}\cdot\hvec{t}$, if $\hvec{t}$ is aligned with the $l$-th segment by the temporal sum pooling.
Using the CAM, we can identify discriminative regions or temporal patterns while considering different class weights for each segment.

\section{Experiments}
\label{sec:exp}
\begin{figure*}[t]
	\centering
	\begin{subfigure}{0.47\linewidth}
	\centering
	\includegraphics[width=\linewidth]{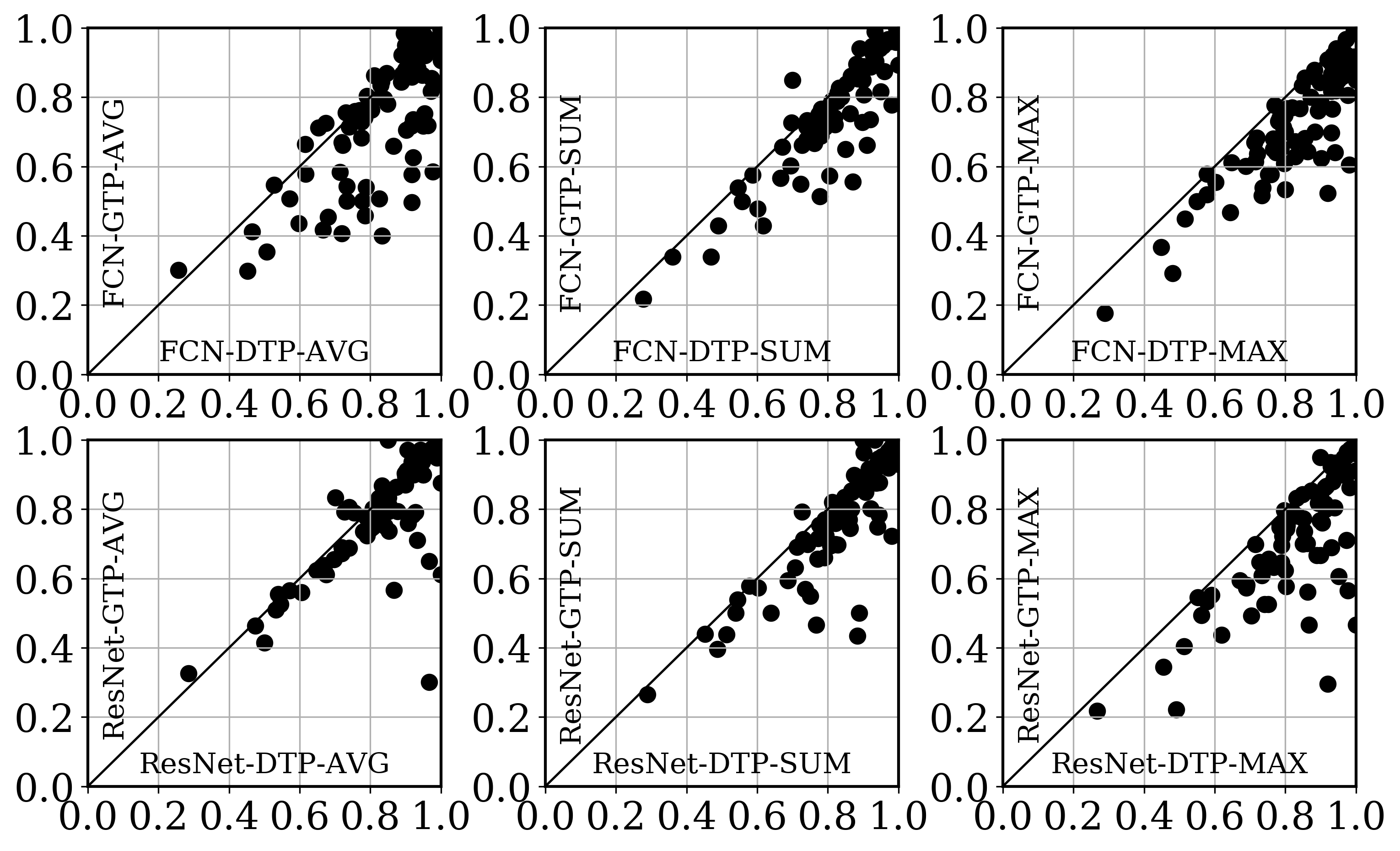}
	\caption{Datasets: UCR univariate time series archive.}
	\end{subfigure}
	\begin{subfigure}{0.47\linewidth}
	\centering
	\includegraphics[width=\linewidth]{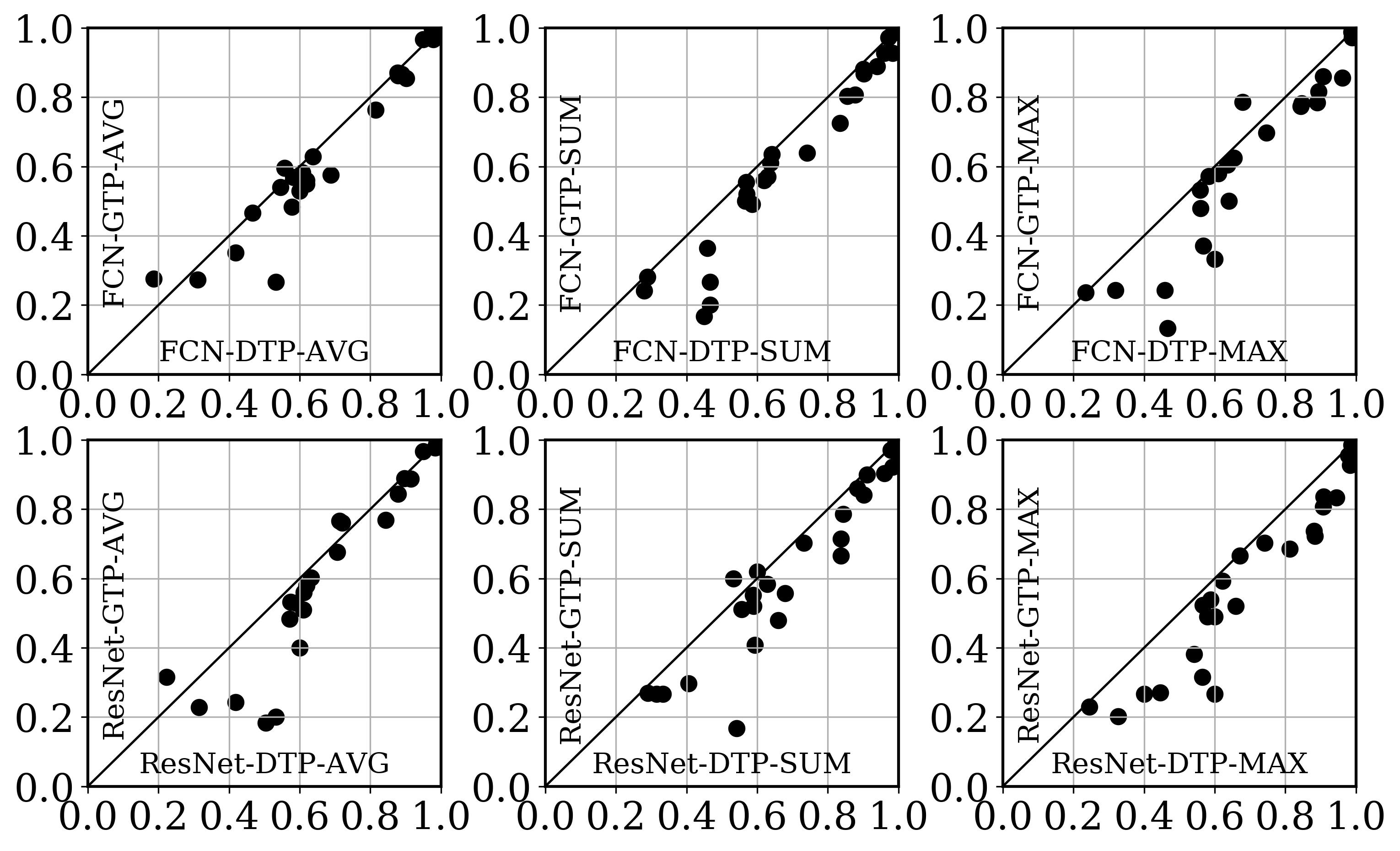}
	\caption{Datasets: UEA multivariate time series archive.}
	\end{subfigure}
	\caption{Comparison of the existing global temporal pooling and the proposed dynamic temporal pooling.}
	\label{fig:comp_dotplot}
\end{figure*}

\subsection{Experimental Settings}
\label{subsec:setting}
\subsubsection{Datasets}
For extensive evaluation, we use 85 univariate time series datasets and 30 multivariate time series datasets from the UCR/UEA repository~\cite{bagnall2018uea,dau2018ucr}.
As the datasets are publicly available as well as collected from a wide range of domains, they have been widely used for the time series classification task.

\subsubsection{Baselines}
As our main baselines, we use CNN classifiers equipped with different temporal pooling layers.
For the experiments, we build three types of temporal pooling with avg, sum, and max operations: global temporal pooling (\gtp), static temporal pooling (\stp) with the fixed pooling size, and the proposed dynamic temporal pooling (\dtp).
We remark that several DNN architectures that have been studied for time series forecasting, for instance, dilated CNN~\cite{oord2016wavenet}, autoregressive model based on recurrent neural networks~\cite{lai2018modeling,rangapuram2018deep}, and transformer~\cite{li2019enhancing}, cannot be directly applied to time series classification (i.e., instance-level label prediction).
All of them need to be customized or tuned to obtain the representation for an input time series instance; in this sense, our \dtp layer can be easily generalized for the architectures as well, but we leave this for future work.

In addition, we compare our classifiers with nearest neighbor classifiers using various distance measures, especially for univariate time series.\footnote{This type of distance measures is not very well defined for multivariate time series.}
Since the purpose of our work is to enhance the discrimination power of a single classifier, we exclude ensemble classifiers~\cite{bagnall2015time,lines2015time,lines2016hive} that require highly intensive computations for exploiting dozens of classifiers together.

\subsubsection{Implementation details}
We employ two CNN architectures, \fcn and \resnet, specifically designed for time series classification~\cite{wang2017time};
they showed the best accuracy among various types of DNNs~\cite{fawaz2019deep}.
We implement all the CNN classifiers using PyTorch, 
and make use of the Numba compiler to compute the forward and backward recursions of the DTW (Algorithm~\ref{alg:softdtw}) in parallel.\footnote{Due to the condition $L \ll T$ as well as parallel DTW computation, the training times of \gtp and \dtp are almost the same.}
In order to eliminate the benefit from hyperparameter tuning, we train our classifiers
without introducing the weight that balances the two losses, and also fix the number of segments $L$ to 4 for \stp and \dtp.\footnote{The \stp and \dtp layers use $L=4$ if it is not explicitly stated.}
Table~\ref{tbl:param} describes the details of hyperparameters used in our experiments.

\subsubsection{Evaluation strategy}
For quantitative evaluation, we conduct the pairwise posthoc analysis~\cite{benavoli2016should} that statistically ranks different classifiers according to their accuracy over multiple datasets, as done in~\cite{fawaz2019deep,yuan2019locally}.
We visualize the results by the critical difference (CD) diagram~\cite{demvsar2006statistical}
which indicates the average rank of each classifier with thick horizontal lines showing a group of classifiers that are not significantly different ($p=0.05$).
For all the datasets, we repeatedly train each classifier three times with different random seeds, and report the median accuracy.

\begin{table}[t]
    \centering
    \resizebox{0.99\linewidth}{!}{
    \begin{tabular}{|c|c|c|c|c|}
    \hline
         Network & \#Conv. & Normalize & Activate & Regularize   \\\hline
         FCN & 3 & BatchNorm & ReLU & None \\
         ResNet & 9 & BatchNorm & ReLU & None \\\hline\hline
         Network & Optimizer & Epochs & Batch size & Learn. rate \\\hline
         FCN & Adam & 500 & 16 & 0.0001 \\
         ResNet & Adam & 500 & 64 & 0.0001 \\
    \hline
    \end{tabular}
    }
    \caption{Hyperparameters for CNN architectures and their optimization. We follow the setting provided by the previous work~\cite{wang2017time,fawaz2019deep}.}
    \label{tbl:param}
\end{table}

\subsection{Comparison of Different Pooling Layers}
\label{subsec:uniresult}

\begin{figure}[t]
	\centering
	\begin{subfigure}{\linewidth}
	\centering
	\includegraphics[width=\linewidth]{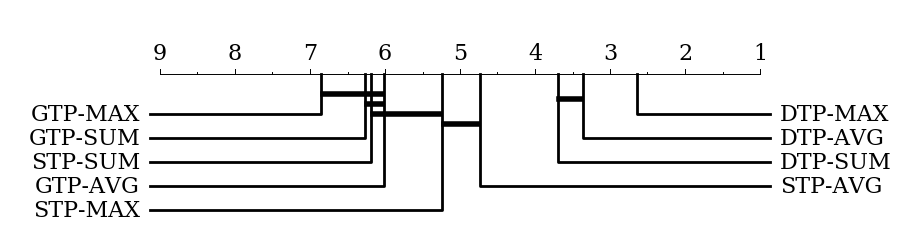}
	\caption{Network: FCN, Datasets: UCR univariate archive.}
	\end{subfigure}
	\begin{subfigure}{\linewidth}
	\centering
	\includegraphics[width=\linewidth]{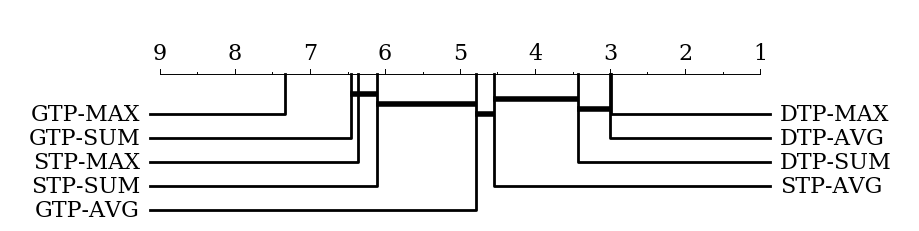}
	\caption{Network: ResNet, Datasets: UCR univariate archive.}
	\end{subfigure}
	\begin{subfigure}{\linewidth}
	\centering
	\includegraphics[width=\linewidth]{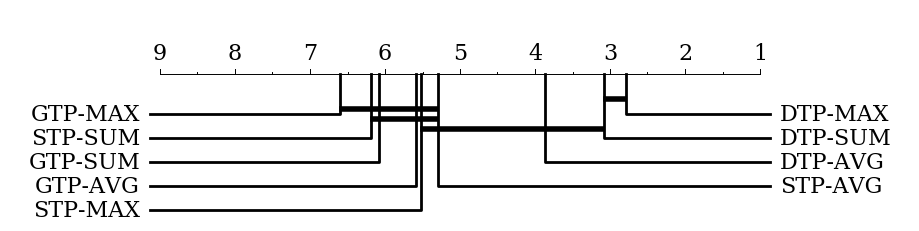}
	\caption{Network: FCN, Datasets: UEA multivariate archive.}
	\end{subfigure}
	\begin{subfigure}{\linewidth}
	\centering
	\includegraphics[width=\linewidth]{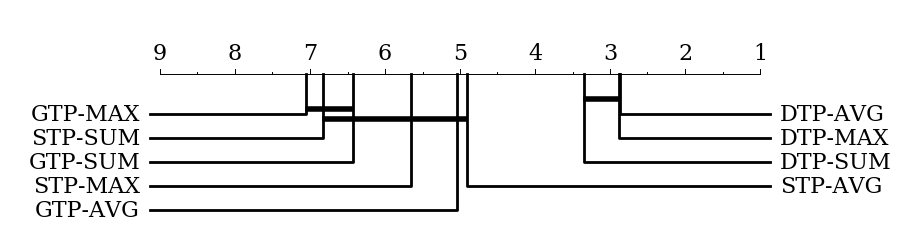}
	\caption{Network: ResNet, Datasets: UEA multivariate archive.}
	\end{subfigure}
	\caption{CD diagrams for comparing CNN classifiers that adopt different types of temporal pooling layers.}
	\label{fig:comp_cd}
\end{figure}

We first directly compare the classification accuracy of our proposed CNN classifier (using \dtp) with that of the baseline CNN classifier (using \gtp). 
In Figure~\ref{fig:comp_dotplot}, a single dot represents each dataset, thus how far each dot is located from the $y=x$ line indicates the performance gap between the two pooling methods.
For all the cases, we observe that most of the datasets are dotted in the lower right side of each figure, showing that \dtp outperforms \gtp regardless of its CNN architectures and pooling operations.
In particular, the performance improvement of \dtp over \gtp becomes larger when it is used with the max operation.
As the max pooling is effective to detect specific features in general, the \dtp-MAX becomes good at discovering such features segment-specifically, which makes the final representation further class-discriminative.

For more statistical evaluation, we also compare different temporal pooling methods (i.e., \gtp, \stp, and \dtp) based on pairwise statistical tests.
Figure~\ref{fig:comp_cd} presents their average rank over a bunch of datasets (85 for the univariate case, and 30 for the multivariate case) with the pairwise statistical differences.
Specifically, \dtp consistently performs the best among all types of temporal pooling while \gtp performs the worst.
In case of \stp, it shows slightly better but not much statistically different performances than \gtp, even though it considers the same number of segments with \dtp;
this implies that pooling the vectors from same-length segments at fixed temporal positions cannot effectively boost the accuracy of CNN classifiers.
On the contrary, our \dtp method, which uses variable-length segments identified by DTW, is capable of modeling high-level features depending on each segment, thus its discrimination power improves a lot.
We can conclude that \dtp successfully utilizes temporal information for its classification compared to \gtp and \stp.

\subsection{Comparison with Nearest Neighbor Classifiers}
\label{subsec:multiresult}

As the nearest neighbor classifier has been one of the powerful benchmarks in the field of univariate time series classification, we compare their performances with ours.
For the nearest neighbor classifier, we consider 8 different distance measures, including DTW, TWE~\cite{marteau2008time}, WDTW~\cite{jeong2011weighted}, MSM~\cite{stefan2012move}, CID-DTW~\cite{batista2014cid}, shapeDTW~\cite{zhao2018shapedtw}, LWDTW~\cite{yuan2019large}, and LSDTW~\cite{yuan2019locally}.

\begin{figure}[t]
	\centering
	\includegraphics[width=\linewidth]{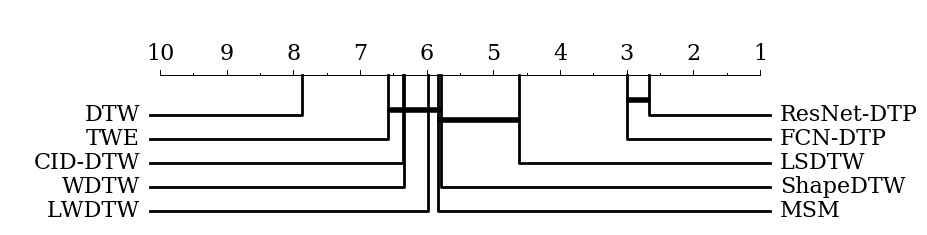}
	\caption{CD diagram for comparing CNN classifiers (\dtp-MAX) with nearest neighbor classifiers.}
	\label{fig:comp_with_knn}
\end{figure}

Figure~\ref{fig:comp_with_knn} clearly shows that the CNN classifiers (\dtp-MAX) beat all the other classifiers by a large margin.
In addition to its higher accuracy, the deep learning approach has some benefits in terms of efficiency and scalability.
To predict the class label of an input time series instance, our classifiers require only a single CNN inference with an additional cost of $O(LT)$ for the DTW alignment between $\pseries$ and $\hseries$.
By contrast, the nearest neighbor classifiers need to perform $O(N)$ comparisons to find the closest instance from the input time series, and each comparison takes $O(T^2)$ for computing the DTW-based distance based on point-to-point matching.
This highly limits the scalability of the nearest neighbor classifiers with respect to $N$ and $T$.

\begin{figure}[t]
	\centering
	\includegraphics[width=\linewidth]{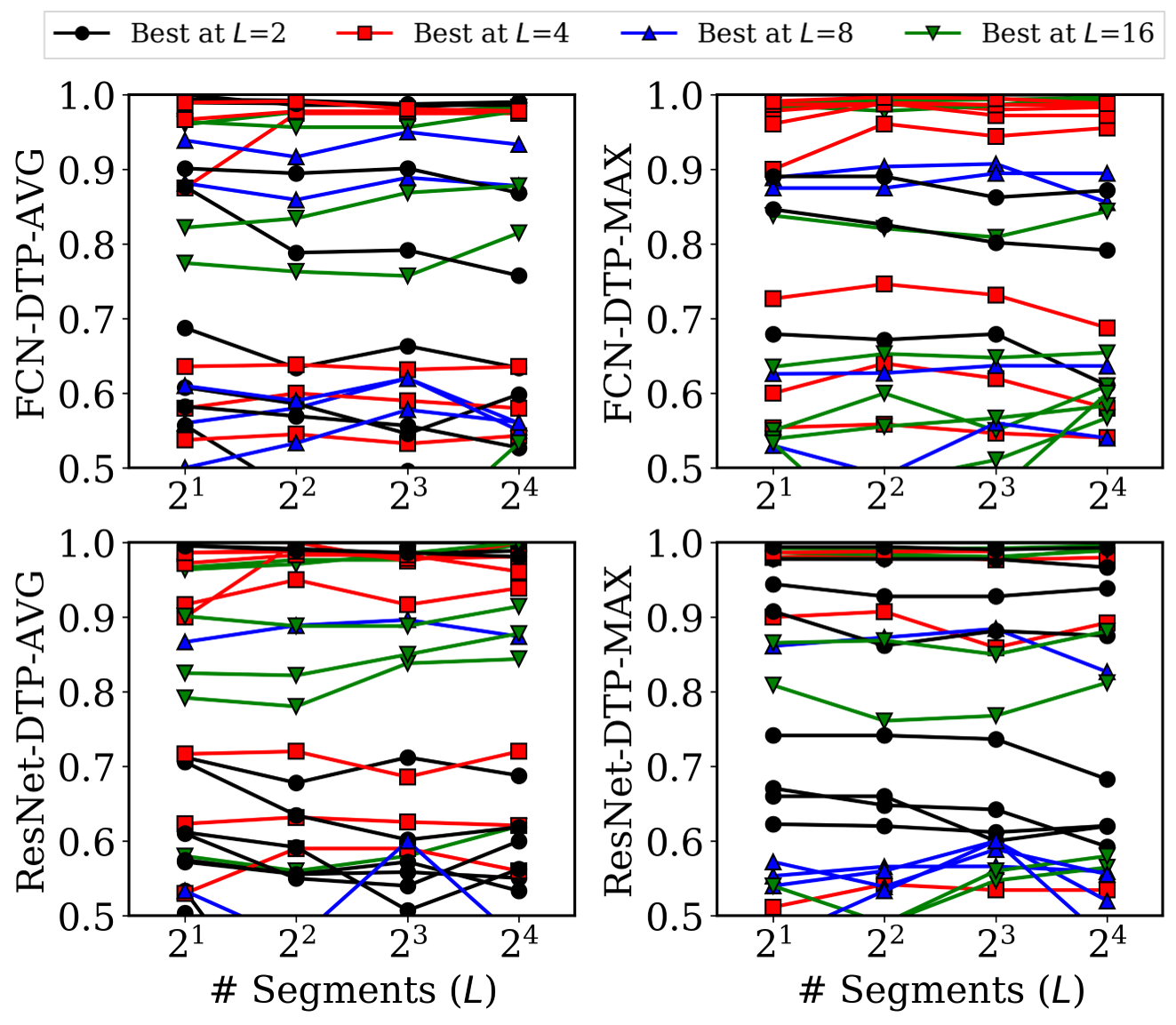}
	\caption{Performance changes of CNN classifiers with the \dtp layer with respect to the number of segments $L$, in terms of classification accuracy.}
	\label{fig:leffect}
\end{figure}

\begin{figure*}[t]
	\centering
	\begin{subfigure}{0.31\linewidth}
	\centering
	\includegraphics[width=0.97\linewidth]{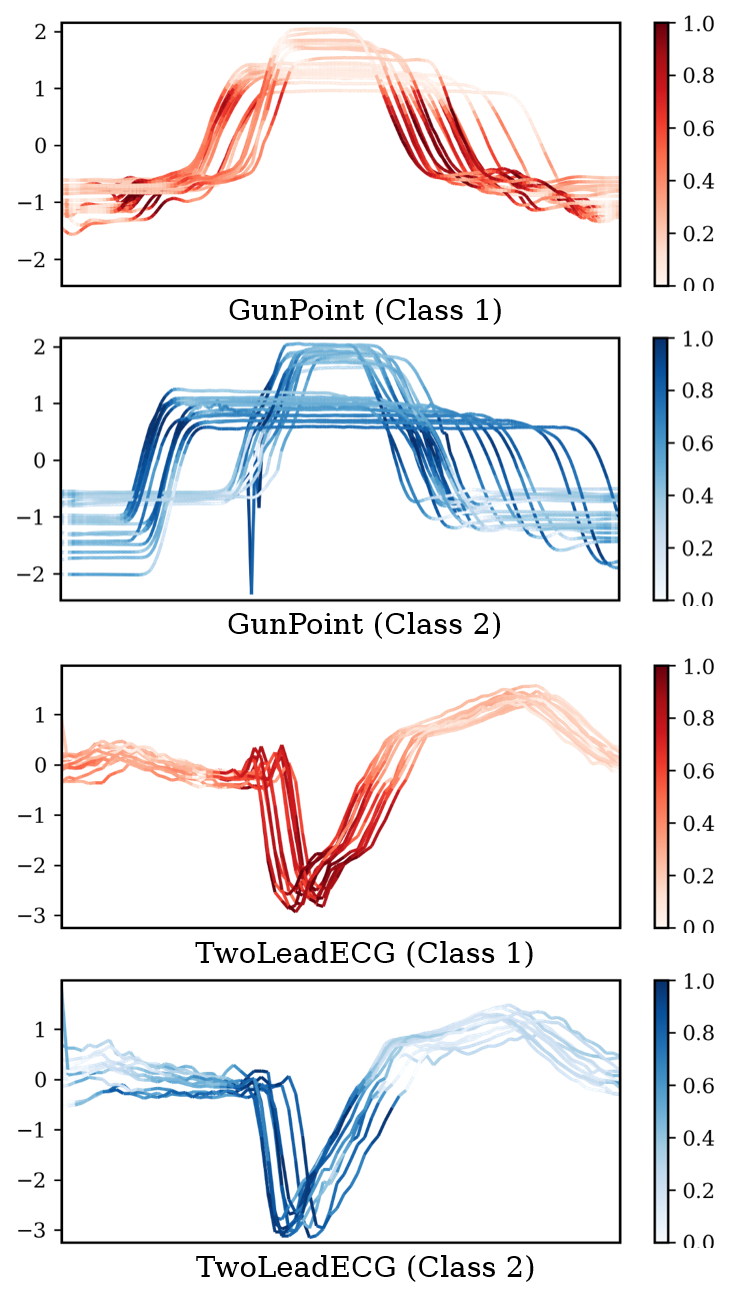}
	\caption{\fcn-\gtp}
	\label{fig:qual_gtp}
	\end{subfigure}
	\begin{subfigure}{0.31\linewidth}
	\centering
	\includegraphics[width=0.97\linewidth]{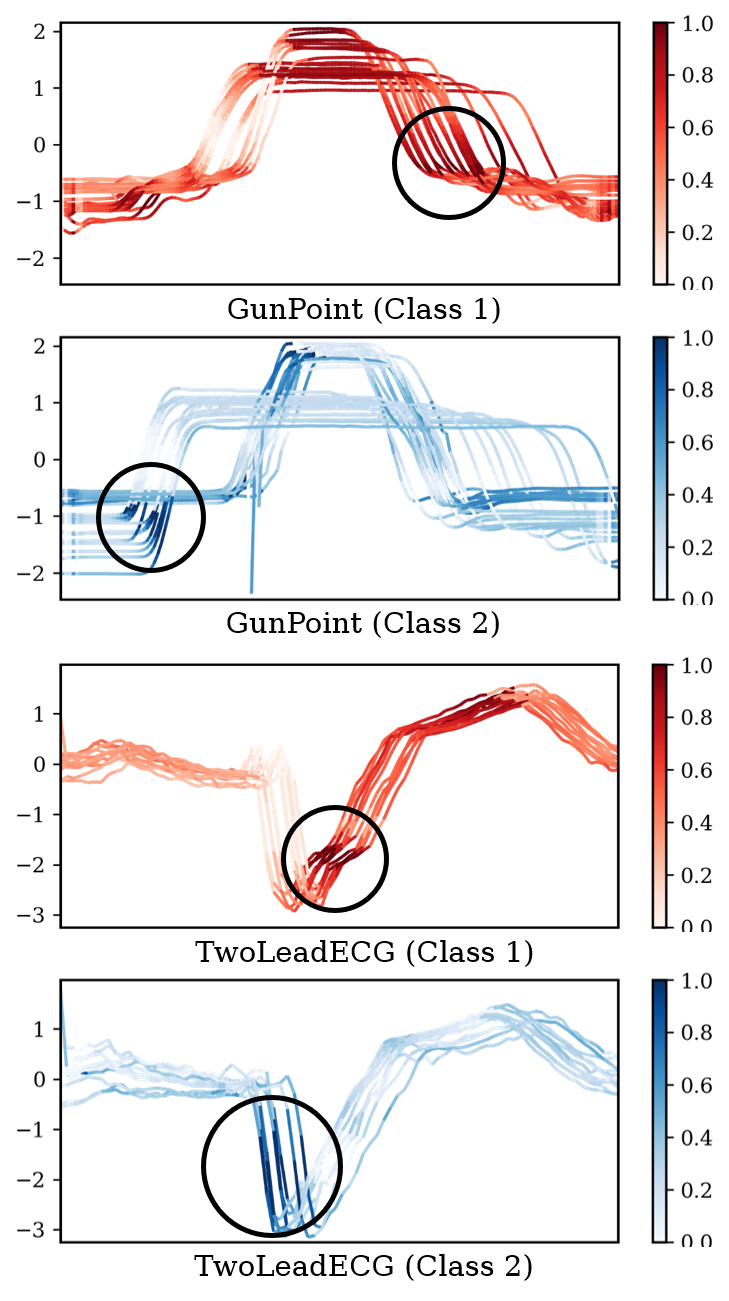}
	\caption{\fcn-\dtp}
	\label{fig:qual_dtp}
	\end{subfigure}
	\begin{subfigure}{0.31\linewidth}
	\centering
	\includegraphics[width=0.95\linewidth]{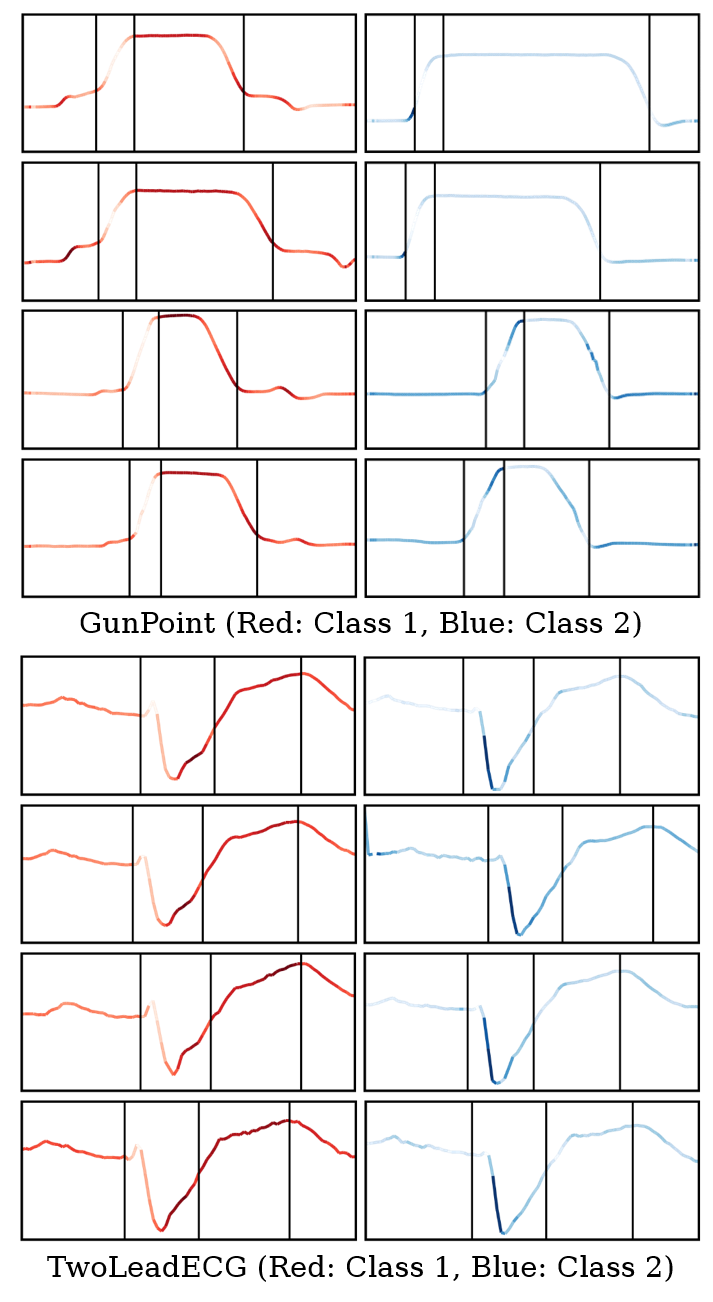}
	\caption{Segmentation results from the \dtp layer}
	\label{fig:segresults}
	\end{subfigure}
	\caption{Time series instances from the GunPoint (Upper) and TwoLeadECG (Lower) datasets, highlighted with their class activation map. Black circles mark the discriminative temporal patterns discovered by our method. (Best viewed in color.)}
	\label{fig:qualresults}
\end{figure*}

\subsection{Parameter Analysis on $L$}
\label{subsec:leffect}
We investigate the performance changes of the CNN classifiers (\dtp-AVG and \dtp-MAX) on the multivariate datasets, increasing the number of segments (i.e., $L$) from $2^1$ to $2^4$.
In Figure~\ref{fig:leffect}, a single line denotes each dataset, and its color is determined by the optimal value of $L$ that achieves the highest accuracy.
The tendency of performance changes is not consistent over most of the datasets, and also the optimal number of segments varies depending on the dataset.
As illustrated in Figure~\ref{fig:patterns}, time series instances in a dataset share their own temporal patterns and lengths, which are distinguished from the ones in other datasets.
For this reason, the optimal number of segments largely depends on such properties of each dataset.
This result strongly indicates that finding the optimal $L$ value for a target dataset can further enhance the performance of the CNN classifiers in practice, compared to the case of using the fixed $L$ in our experiments.

\subsection{Qualitative Analysis}
\label{subsec:qualanal}
To qualitatively compare the localization performance of \gtp and \dtp, we visualize their CAM scores by highlighting the input time series proportionally to the scores.
In Figure~\ref{fig:qual_gtp} and \ref{fig:qual_dtp}, the time series instances of class 1 and 2 are colored in red and blue, respectively, and 
we use the CAM scores after normalizing them in the range of $[0, 1]$ for each time series instance.
Although the two CNN classifiers achieve almost the same accuracy for both the datasets (i.e., GunPoint and TwoLeadECG), they highlight different regions as the discriminative temporal patterns that most contribute to predicting its class label.
It is worth noting that the localized regions discovered by \dtp are more clearly distinguishable between the classes compared to \gtp, which results in better interpretability of the CNN classifiers.

Furthermore, Figure~\ref{fig:segresults} provides segmentation results obtained from the \dtp layer; 
each time series instance is divided into four segments by vertical lines.
The $l$-th segments ($l=1,\ldots,4$) of all instances share similar (or consistent) temporal patterns, even where their original input series are not temporally aligned.
The results support that our prototypical hidden series $\pseries$ successfully encodes the prototypical high-level features of the segments in a temporal order, thereby the \dtp layer can perform semantic segmentation based on the DTW alignment between $\pseries$ and $\hseries$.


\section{Conclusion}
\label{sec:concl}
This paper proposes a dynamic temporal pooling, termed as \dtp, which outputs the pooled vectors from temporally-ordered segments;
this enables CNN classifiers to make use of the segment-level fully-connected layer for time series classification.
We present a learning framework to simultaneously optimize the network parameters of a CNN classifier and the prototypical hidden series that encodes the latent semantic of the segments.
By finding the optimal alignment between the prototypical hidden series and the hidden representation of a target time series, the \dtp layer is able to partition the whole series into a fixed number of segments.
Our extensive experiments show that the proposed \dtp layer outperforms other baseline pooling layers, and it successfully identifies consistent segments from time series instances that are out of temporal alignment.

\section{Acknowledgements}
This work was supported by the NRF grant funded by the MSIT (No.~2020R1A2B5B03097210), and the IITP grant funded by the MSIT (No.~2018-0-00584, 2019-0-01906).

\bibliographystyle{aaai21}
\bibliography{BIB/dtw-pool}

\end{document}